\documentclass{article}
\usepackage{ijcai19}

\usepackage{times}
\usepackage{soul}
\usepackage{url}
\usepackage[hidelinks]{hyperref}
\usepackage[utf8]{inputenc}
\usepackage[small]{caption}
\usepackage{graphicx}
\usepackage{amsmath}
\usepackage{booktabs}
\usepackage{algorithm}
\usepackage{algorithmic}
\title{Saliency Maps Generation for Automatic Text Summarization}

\author{
David Tuckey
\and
Krysia Broda
\and
Alessandra Russo
\affiliations
Department of Computing, Imperial College London\\
\emails
\{david.tuckey17, k.broda, a.russo\}@imperial.ac.uk
}

\begin{document}

\maketitle

\begin{abstract}
    Saliency map generation techniques are at the forefront of explainable AI literature for a broad range of machine learning applications. Our goal is to question the limits of these approaches on more complex tasks. In this paper we apply Layer-Wise Relevance Propagation (LRP) to a sequence-to-sequence attention model trained on a text summarization dataset. We obtain unexpected saliency maps and discuss the rightfulness of these ``explanations". We argue that we need a quantitative way of testing the counterfactual case to judge the truthfulness of the saliency maps. We suggest a protocol to check the validity of the importance attributed to the input and show that the saliency maps obtained sometimes capture the real use of the input features by the network, and sometimes do not. We use this example to discuss how careful we need to be when accepting them as explanation. 
\end{abstract}

\section{Introduction}

Ever since the LIME algorithm \cite{Ribeiro2016}, "explanation" techniques focusing on finding the importance of input features in regard of a specific prediction have soared and we now have many ways of finding saliency maps (also called heat-maps because of the way we like to visualize them). We are interested in this paper by the use of such a technique in an extreme task that highlights questions about the validity and evaluation of the approach. We would like to first set the vocabulary we will use. We agree that saliency maps are not explanations in themselves and that they are more similar to attribution, which is only one part of the human explanation process \cite{Miller2017}. We will prefer to call this importance mapping of the input an \textit{attribution} rather than an explanation. We will talk about the importance of the input relevance score in regard to the model's computation and not make allusion to any human understanding of the model as a result. 

There exist multiple ways to generate saliency maps over the input for non-linear classifiers \cite{Bach2015,Montavon2017,Samek2017}. We refer the reader to \cite{Adadi2018} for a survey of explainable AI in general. We use in this paper Layer-Wise Relevance Propagation (LRP) \cite{Bach2015} which aims at redistributing the value of the classifying function on the input to obtain the importance attribution. It was first created to ``explain" the classification of neural networks on image recognition tasks. It was later successfully applied to text using convolutional neural networks (CNN) \cite{Arras2017a} and then Long-Short Term Memory (LSTM) networks for sentiment analysis \cite{Arras2017}. 

Our goal in this paper is to test the limits of the use of such a technique for more complex tasks, where the notion of input importance might not be as simple as in topic classification or sentiment analysis. We changed from a classification task to a generative task and chose a more complex one than text translation (in which we can easily find a word to word correspondence/importance between input and output). We chose text summarization. We consider abstractive and informative text summarization, meaning that we write a summary ``in our own words" and retain the important information of the original text. We refer the reader to \cite{Radev2002} for more details on the task and the different variants that exist. Since the success of deep sequence-to-sequence models for text translation \cite{Bahdanau2014}, the same approaches have been applied to text summarization tasks \cite{Rush2015,See2017,Nallapati2016} which use architectures on which we can apply LRP.

We obtain one saliency map for each word in the generated summaries, supposed to represent the use of the input features for each element of the output sequence. We observe that all the saliency maps for a text are nearly identical and decorrelated with the attention distribution. We propose a way to check their validity by creating what could be seen as a counterfactual experiment from a synthesis of the saliency maps, using the same technique as in Arras et al. \shortcite{Arras2017}. We show that in some but not all cases they help identify the important input features and that we need to rigorously check importance attributions before trusting them, regardless of whether or not the mapping ``makes sense" to us. We finally argue that in the process of identifying the important input features, verifying the saliency maps is as important as the generation step, if not more.

\section{The Task and the Model}

We present in this section the baseline model from See et al. \shortcite{See2017} trained on the CNN/Daily Mail dataset. We reproduce the results from See et al. \shortcite{See2017} to then apply LRP on it. 

\subsection{Dataset and Training Task}

 The CNN/Daily mail dataset \cite{Nallapati2016} is a text summarization dataset adapted from the Deepmind question-answering dataset \cite{Hermann2015}. It contains around three hundred thousand news articles coupled with summaries of about three sentences. These summaries are in fact ``highlights" of the articles provided by the media themselves. Articles have an average length of 780 words and the summaries of 50 words. We had 287 000 training pairs and 11 500 test pairs. Similarly to See et al. \shortcite{See2017}, we limit during training and prediction the input text to 400 words and generate summaries of 200 words. We pad the shorter texts using an UNKNOWN token and truncate the longer texts.
 We embed the texts and summaries using a vocabulary of size 50 000, thus recreating the same parameters as See et al. \shortcite{See2017}.
 
 \subsection{The Model}
 
 The baseline model is a deep sequence-to-sequence encoder/decoder model with attention. The encoder is a bidirectional Long-Short Term Memory(LSTM) cell \cite{Hochreiter1997} and the decoder a single LSTM cell with attention mechanism. The attention mechanism is computed as in \cite{Bahdanau2014} and we use a greedy search for decoding. 
 We train end-to-end including the words embeddings. The embedding size used is of 128 and the hidden state size of the LSTM cells is of 254. 
 
 \subsection{Obtained Summaries}
 
 We train the 21 350 992 parameters of the network for about 60 epochs until we achieve results that are qualitatively equivalent to the results of See et al. \shortcite{See2017}. We obtain summaries that are broadly relevant to the text but do not match the target summaries very well. We observe the same problems such as wrong reproduction of factual details, replacing rare words with more common alternatives or repeating non-sense after the third sentence. We can see in Figure \ref{fig:example_summary} an example of summary obtained compared to the target one.
 
 \begin{figure}[h!]
     \centering
     \framebox{\parbox{\dimexpr\linewidth-2\fboxsep-2\fboxrule}{\textbf{Target summary} : marseille prosecutor says ``so far no videos were used in the crash
     investigation" despite media reports. journalists at bild and paris
     match are "very confident" the video clip is real, an editor says.
     andreas lubitz had informed his lufthansa training school of an episode
     of severe depression, airline says.
     }}
     \framebox{\parbox{\dimexpr\linewidth-2\fboxsep-2\fboxrule}{\textbf{Generated summary} : $<$s$>$ the $<$UNK$>$ was found in a crash on the board flight . . the video was found by a source close to the investigation . . the video was found by a source close to the investigation ...[truncated]
     }}
     \caption{Top : example of target generated. Bottom : generated summary for the same text}
     \label{fig:example_summary}
 \end{figure}

\begin{figure*}[t]
    \centering
    \includegraphics[width=0.7\linewidth]{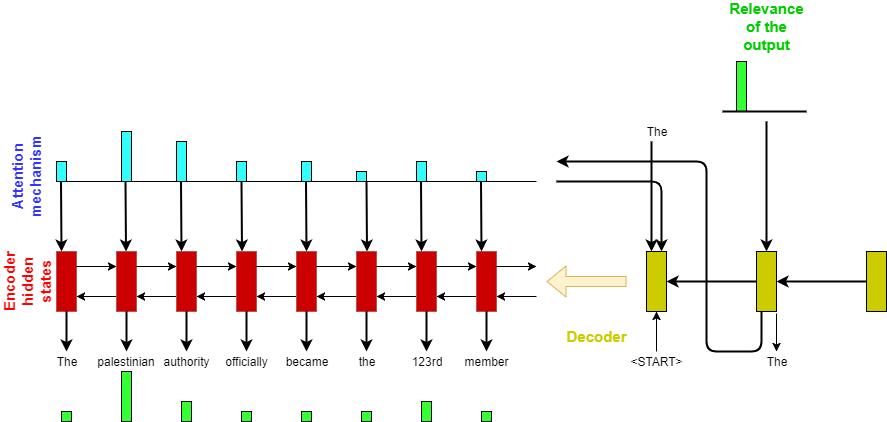}
    \caption{Representation of the propagation of the relevance from the output to the input. It passes through the decoder and attention mechanism for each previous decoding time-step, then is passed onto the encoder which takes into account the relevance transiting in both direction due to the bidirectional nature of the encoding LSTM cell.}
    \label{fig:relevance_transmission}
\end{figure*}
 
 The ``summaries" we generate are far from being valid summaries of the information in the texts but are sufficient to look at the attribution that LRP will give us. They pick up the general subject of the original text. 
 
 \section{Layer-Wise Relevance Propagation}
 
 We present in this section the Layer-Wise Relevance Propagation (LRP) \cite{Bach2015} technique that we used to attribute importance to the input features, together with how we adapted it to our model and how we generated the saliency maps. LRP redistributes the output of the model from the output layer to the input by transmitting information backwards through the layers. We call this propagated backwards importance the \textit{relevance}. LRP has the particularity to attribute negative and positive relevance: a positive relevance is supposed to represent evidence that led to the classifier's result while negative relevance represents evidence that participated negatively in the prediction. 
 
 \subsection{Mathematical Description}
 
 We initialize the relevance of the output layer to the value of the predicted class before softmax and we then describe locally the propagation backwards of the relevance from layer to layer.
 For normal neural network layers we use the form of LRP with epsilon stabilizer \cite{Bach2015}. We write down $R_{i\leftarrow j}^{(l, l+1)}$ the relevance received by the neuron $i$ of layer $l$ from the neuron $j$ of layer $l+1$: 
 
\begin{equation}
\begin{split}
    \label{eq:final_LRP_formula}
    R_{i\leftarrow j}^{(l, l+1)} &= \dfrac{w_{i\rightarrow j}^{l,l+1}\textbf{z}^l_i + \dfrac{\epsilon \textrm{ sign}(\textbf{z}^{l+1}_j) + \textbf{b}^{l+1}_j}{D_l}}{\textbf{z}^{l+1}_j + \epsilon * \textrm{ sign}(\textbf{z}^{l+1}_j)} * R_j^{l+1} \\
\end{split}
\end{equation}

where $w_{i\rightarrow j}^{l,l+1}$ is the network's weight parameter set during training, $\textbf{b}^{l+1}_j$ is the bias for neuron $j$ of layer $l+1$, $\textbf{z}^{l}_i$ is the activation of neuron $i$ on layer $l$, $\epsilon$ is the stabilizing term set to 0.00001 and $D_l$ is the dimension of the $l$-th layer.

The relevance of a neuron is then computed as the sum of the relevance he received from the above layer(s).\\

For LSTM cells we use the method from Arras et al.\shortcite{Arras2017} to solve the problem posed by the element-wise multiplications of vectors. Arras et al. noted that when such computation happened inside an LSTM cell, it always involved a ``gate" vector and another vector containing information. The gate vector containing only value between 0 and 1 is essentially filtering the second vector to allow the passing of ``relevant" information. Considering this, when we propagate relevance through an element-wise multiplication operation, we give all the upper-layer's relevance to the ``information" vector and none to the ``gate" vector. 

\subsection{Generation of the Saliency Maps}

We use the same method to transmit relevance through the attention mechanism back to the encoder because Bahdanau's attention \cite{Bahdanau2014} uses element-wise multiplications as well. We depict in Figure \ref{fig:relevance_transmission} the transmission end-to-end from the output layer to the input through the decoder, attention mechanism and then the bidirectional encoder. We then sum up the relevance on the word embedding to get the token's relevance as Arras et al. \shortcite{Arras2017}. 

The way we generate saliency maps differs a bit from the usual context in which LRP is used as we essentially don't have one classification, but 200 (one for each word in the summary). We generate a relevance attribution for the 50 first words of the generated summary as after this point they often repeat themselves.

This means that for each text we obtain 50 different saliency maps, each one supposed to represent the relevance of the input for a specific generated word in the summary. 

\section{Experimental results}

In this section, we present our results from extracting attributions from the sequence-to-sequence model trained for abstractive text summarization. We first have to discuss the difference between the 50 different saliency maps we obtain and then we propose a protocol to validate the mappings.

\subsection{First Observations}

\begin{figure}[h]
    \centering
    \includegraphics[width=0.49\linewidth]{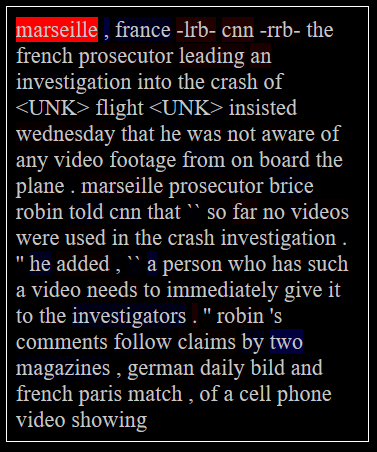}
    \includegraphics[width=0.49\linewidth]{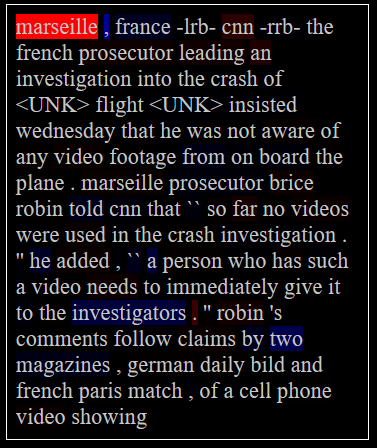}
    \caption{Left : Saliency map over the truncated input text for the second generated word ``the". Right : Saliency map over the truncated input text for the 25th generated word ``investigation". We see that the difference between the mappings is marginal.}
    \label{fig:comparison_heatmaps}
\end{figure}

The first observation that is made is that for one text, the 50 saliency maps are almost identical. Indeed each mapping highlights mainly the same input words with only slight variations of importance. We can see in Figure \ref{fig:comparison_heatmaps} an example of two nearly identical attributions for two distant and unrelated words of the summary. The saliency map generated using LRP is also uncorrelated with the attention distribution that participated in the generation of the output word. The attention distribution changes drastically between the words in the generated summary while not impacting significantly the attribution over the input text. We deleted in an experiment the relevance propagated through the attention mechanism to the encoder and didn't observe much changes in the saliency map. 

It can be seen as evidence that using the attention distribution as an ``explanation" of the prediction can be misleading. It is not the only information received by the decoder and the importance it ``allocates" to this attention state might be very low. What seems to happen in this application is that most of the information used is transmitted from the encoder to the decoder and the attention mechanism at each decoding step just changes marginally how it is used. Quantifying the difference between attention distribution and saliency map across multiple tasks is a possible future work. 

The second observation we can make is that the saliency map doesn't seem to highlight the right things in the input for the summary it generates. The saliency maps on Figure \ref{fig:comparison_heatmaps} correspond to the summary from Figure \ref{fig:example_summary}, and we don't see the word ``video" highlighted in the input text, which seems to be important for the output. 

This allows us to question how good the saliency maps are in the sense that we question how well they actually represent the network's use of the input features. We will call that \textit{truthfulness of the attribution} in regard to the computation, meaning that an attribution is truthful in regard to the computation if it actually highlights the important input features that the network attended to during prediction. We proceed to measure the truthfulness of the attributions by validating them quantitatively.

\subsection{Validating the Attributions}

We propose to validate the saliency maps in a similar way as Arras et al. \shortcite{Arras2017} by incrementally deleting ``important" words from the input text and observe the change in the resulting generated summaries. 

We first define what ``important" (and ``unimportant") input words mean across the 50 saliency maps per texts. Relevance transmitted by LRP being positive or negative, we average the absolute value of the relevance across the saliency maps to obtain one ranking of the most ``relevant" words. The idea is that input words with negative relevance have an impact on the resulting generated word, even if it is not participating positively, while a word with a relevance close to zero should not be important at all. We did however also try with different methods, like averaging the raw relevance or averaging a scaled absolute value where negative relevance is scaled down by a constant factor. The absolute value average seemed to deliver the best results. 

We delete incrementally the important words (words with the highest average) in the input and compared it to the control experiment that consists of deleting the least important word and compare the degradation of the resulting summaries. We obtain mitigated results: for some texts, we observe a quick degradation when deleting important words which are not observed when deleting unimportant words (see Figure \ref{fig:good_experiment}), but for other test examples we don't observe a significant difference between the two settings (see Figure \ref{fig:bad_experiment}).

 \begin{figure}[h!]
     \centering
     \framebox{\parbox{\dimexpr\linewidth-2\fboxsep-2\fboxrule}{\textbf{Deleting 7\% most important words} : the $<$UNK$>$ $<$UNK$>$ $<$UNK$>$ $<$UNK$>$ $<$UNK$>$ $<$UNK$>$ $<$UNK$>$ $<$UNK$>$ $<$UNK$>$ $<$UNK$>$ $<$UNK$>$ $<$UNK$>$ . . $<$UNK$>$ $<$UNK$>$ $<$UNK$>$ $<$UNK$>$[truncated]
     }}
     \framebox{\parbox{\dimexpr\linewidth-2\fboxsep-2\fboxrule}{\textbf{Deleting 7\% least important words} : the $<$UNK$>$ was filmed by the magazine and the $<$UNK$>$. the video was found by a source close to the investigation. the $<$UNK$>$ said the video was recovered from a phone at the wreckage site [truncated]
     }}
     \caption{Summary from Figure \ref{fig:example_summary} generated after deleting important and unimportant words from the input text. We observe a significant difference in summary degradation between the two experiments, where the decoder just repeats the UNKNOWN token over and over.}
     \label{fig:good_experiment}
 \end{figure}

 \begin{figure}[h!]
     \centering
     \framebox{\parbox{\dimexpr\linewidth-2\fboxsep-2\fboxrule}{\textbf{Deleting 7\% most important words} : the $<$UNK$>$ mass index was carried out against the taliban in 2012 . . the $<$UNK$>$ mass index was part of china 's ` strike hard ' campaign against the notion that the mid-1970s was killed[truncated]
     }}
     \framebox{\parbox{\dimexpr\linewidth-2\fboxsep-2\fboxrule}{\textbf{Deleting 7\% least important words} : the $<$UNK$>$ mass index was carried out in the wake of the horrific attack on a school in peshawar . . the government has issued a ban on executions in the country[truncated]
     }}
     \caption{Summary from another test text generated after deleting important and unimportant words from the input text. We observe less significant difference in summary degradation between the two experiments.}
     \label{fig:bad_experiment}
 \end{figure}

One might argue that the second summary in Figure \ref{fig:bad_experiment} is better than the first one as it makes better sentences but as the model generates inaccurate summaries, we do not wish to make such a statement.

This however allows us to say that the attribution generated for the text at the origin of the summaries in Figure \ref{fig:good_experiment} are truthful in regard to the network's computation and we may use it for further studies of the example, whereas for the text at the origin of Figure \ref{fig:bad_experiment} we shouldn't draw any further conclusions from the attribution generated. 

One interesting point is that one saliency map didn't look ``better" than the other, meaning that there is no apparent way of determining their truthfulness in regard of the computation without doing a quantitative validation. This brings us to believe that even in simpler tasks, the saliency maps might make sense to us (for example highlighting the animal in an image classification task), without actually representing what the network really attended too, or in what way. 

We defined without saying it the counterfactual case in our experiment: ``Would the important words in the input be deleted, we would have a different summary". Such counterfactuals are however more difficult to define for image classification for example, where it could be applying a mask over an image, or just filtering a colour or a pattern. We believe that defining a counterfactual and testing it allows us to measure and evaluate the truthfulness of the attributions and thus weight how much we can trust them.

\section{Conclusion}

In this work, we have implemented and applied LRP to a sequence-to-sequence model trained on a more complex task than usual: text summarization. We used previous work to solve the difficulties posed by LRP in LSTM cells and adapted the same technique for Bahdanau et al. \shortcite{Bahdanau2014} attention mechanism. 

We observed a peculiar behaviour of the saliency maps for the words in the output summary: they are almost all identical and seem uncorrelated with the attention distribution. We then proceeded to validate our attributions by averaging the absolute value of the relevance across the saliency maps. We obtain a ranking of the word from the most important to the least important and proceeded to delete one or another. 

We showed that in some cases the saliency maps are truthful to the network's computation, meaning that they do highlight the input features that the network focused on. But we also showed that in some cases the saliency maps seem to not capture the important input features. This brought us to discuss the fact that these attributions are not sufficient by themselves, and that we need to define the counter-factual case and test it to measure how truthful the saliency maps are. 

Future work would look into the saliency maps generated by applying LRP to pointer-generator networks and compare to our current results as well as mathematically justifying the average that we did when validating our saliency maps. Some additional work is also needed on the validation of the saliency maps with counterfactual tests. The exploitation and evaluation of saliency map are a very important step and should not be overlooked.

\newpage
\bibliographystyle{named}
\bibliography{library}

\end{document}